# Opinion Extraction as A Structured Sentiment Analysis using Transformers

Tian Zhu, Yucheng Liu
W 266: Natural Language Processing
UC Berkeley School of Information
tian_zhu@berkeley.edu  yusufliu@berkeley.edu

## Abstract

Relationship extraction and named entity recognition have always been considered as two distinct tasks that require different input data, labels, and models. However, both are essential for structured sentiment analysis. We believe that both tasks can be combined into a single stacked model with the same input data. We performed different experiments to find the best model to extract multiple opinion tuples from a single sentence. The opinion tuples will consist of holders, targets, and expressions. With the opinion tuples, we will be able to extract the relationship we need.

## 1 Introduction

People's opinion is always the most valuable data that we want to collect. Based on their opinions, a company can easily predict the sales of a product or the results of an election. There are many challenges to creating a system that can process subjective information effectively. Natural language processing (NLP) is an artificial intelligence that specializes in analyzing human language especially at the text level, which is currently the best foundation for building an opinion extraction system.

Today, most of the existing NLP approaches are still based on the syntactic representation of text, a method that relies mainly on word co-occurrence frequencies. Such algorithms are limited by the fact that they can process only the information that they can 'see'. (E. Cambria and B. White, 2014) As we know, NER, named entity recognition, is the basic building block of opinion extraction. Any opinion will contain a target and an expression. So, identifying specific words or phrases ('entities') and categorizing them from text, will be the first challenge we have to solve to extract the opinion. For example, persons, locations, diseases, genes, or medication are all possible targets for one to express opinions on.

A common NER task is mapping named entities to concepts in a vocabulary, which in our use case is identifying the possible targets of an opinion. For example. In the opinion sentence "I love school", school is the target and love is the expression. This task often leverages shallow parsing for candidate entities (e.g., the noun phrase 'chest tenderness'); however, sometimes the opinions are divided across multiple phrases (Prakash M,2011), and sometimes multiple opinions can be found in a single sentence. Thus, our goal is to develop a strategy for extracting multiple opinions from text with a clear definition of source, target, and expression.

## 2 Background

The problem we are trying to solve is extracting all opinion tuples in the text with a single NLP pipeline and assigning sentiment to each of the opinion tuples. For each of our opinion tuples, we will include holder (h), target (t), a polarity (p), and a sentiment expression (e). Those four components are essential to fully resolve the sentiment analysis problem. (Liu, 2012) Most of the current NLP work splits the problem into subproblems like sentiment detection, and named entity recognition, which avoids performing the full task, or on simplified and idealized tasks.

Based on (Barnes et al., 2021), the division of structured sentiment into sub-tasks is counterproductive, because they barely see any experiments suggesting that the addition to the pipeline improves the overall resolution of sentiment, or do not consider the inter-dependencies of the various sub-tasks. As such, they proposed a unified approach to the structured sentiment which jointly predicts all elements of an opinion tuple and their relations.

However, we found that there has been a lot of work done on structured sentiment analysis. (Choi et al., 2006; Yang and Cardie, 2012) used CRFS with the named-entity tagger, sentiment lexicons, and dependency parsers, which result in a very solid baseline. Also, on the MPQA dataset (Wiebe et al., 2005), it follows the standard procedure of extracting the holders, targets, and expressions, then assigning polarity to each term, and predicting their relationships.

Inspired by (Choi et al., 2006; Yang and Cardie, 2012), we believe that there is still value in the division of structured sentiment into sub-tasks because we experimented on splitting the task into pipelines, which improves the overall resolution of sentiment. As such, we propose a two-stage approach, which first uses the NER technique to extract source, target, holder, then extracts relationships from those NER tuples to the structured sentiment which jointly predicts all elements of an opinion tuple and their relations.

Figure 1 Structured Sentiment Graph Example

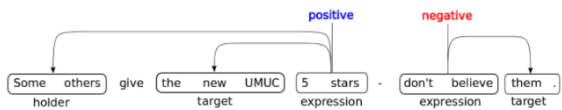

## 3 Methods

### 3.1 Dataset

The goal of the experiment is solely focused on opinion extraction in the English language. For a potential opinion, we wanted to at least have the labels for the holder, target, and polar expression available.

There are three datasets we've acquired that match our criteria. Thus, the experiments are also performed on the three datasets. The **MPQA** dataset is a labeled English dataset on news in Sri Lanka. The **DSUNI** is a labeled English dataset on reviews of online universities. The **OpenNER** dataset consists of labeled reviews of hotels from the guests. We expect that the first dataset would cover the political context while the latter two would cover the opinion towards commercial and non-commercial entities.

Although the datasets contain the polarity information, we did not use or intend to use the polarity information for the task of opinion extraction. The polarity of opinion is often context-driven and attempting to cover such scope with small datasets often only introduces noise. For example, "the hotel elevator was not fast" can have different polarities depending on whether the person preferred fast or slow elevators. As a result, we decided to exclude polarity from the experiment and focus only on the task of opinion extraction.

After creating the data distribution table (Table 1), we found that the dataset is extremely unbalanced, we see there exists only a small number of sentences containing opinion (source, target, or expression). To get the exact number, we mapped the data into different groups by counting how many distinct labels exist within each sentence. After processing the data, we found that there are 6380 sentences with only "O" labels, and 1929 sentences with 3 different labels, and 1055 sentences with all four labels (O, Source, Target, Expression). Thus, we decided to up-sample the dataset to create balanced data. While training the data, we also found that model weird on labels with target and expression sentences. Thus, we manually read through those types of sentences and found that there is a labeling overlap issue for those types of sentences, which result in misprediction in our model, so in the data processing stage we added a filter for those sentences.

Table 1 Data Distribution of the Training Data

| Dataset | total_sentence | source_count | source_max_count | source_avg_count | target_count | target_max_count | target_avg_count | exp_count | exp_max_count | exp_avg_count |
|---|---|---|---|---|---|---|---|---|---|---|
| Darmstadt_unis | 2230.0 | 55.0 | 1.0 | 0.02 | 707.0 | 4.0 | 0.32 | 770.0 | 5.0 | 0.35 |
| MPQA | 5628.0 | 1048.0 | 3.0 | 0.19 | 1293.0 | 5.0 | 0.23 | 1473.0 | 5.0 | 0.26 |
| OpeNER_en | 1640.0 | 168.0 | 3.0 | 0.10 | 1951.0 | 16.0 | 1.19 | 2455.0 | 13.0 | 1.50 |

### 3.2 Proposed Model

The proposed model considers the task of opinion extraction as a three-part task
- **opinion component extraction**
- **component relationship recognition**
- **structured sentiment graph aggregation**

The opinion component extraction step will first extract the corresponding holder, target, and the expression in the text while the component relationship recognition model identifies which pairs of holder/target/expression have a relationship. In the end, the aggregation model will combine the extracted information about the components and their relationships, creating the final representation of opinion.

The **opinion component extraction** piece can be seen as a NER task. In this case, the dataset was converted to a CoNLL format where the labels of the holder, target, and the expression were applied to the corresponding text for training. We used a pre-trained RoBERTa model and fine-tuned it on the downstream NER task. The specific pre-trained version used was the roberta-base model.

The **component relationship recognition** task was treated as a binary classification task. In this step, we transformed the original dataset again to obtain context (i.e., the full original text) and entity/expression pairs, along with the labels of whether there exists a relationship between the pair. The context and the entity/expression pair are then tokenized using the pre-trained RoBERTa tokenizer and concatenated into one single embedding. The embedding vector generated is passed into a transformer architecture for training.

The third task of **structured sentiment graph aggregation** combines the result of the previous two tasks and creates the final structured sentiment graph. Once the corresponding entities of holder, target, and expressions are extracted, each target/holder is combined with the context and the expression and fed into the relationship model. The aggregation model then uses the binary relationship and the extracted entities to form the structured sentiment graph desired by the output.

### 3.3 Baseline Model

For the baseline model, we implemented a most common class predictor. In this use case, based on the label distribution, the most common tag was "O", which means this word is not part of an opinion. For our NER task, when we predict everything to be "O", we will expect a high precision for our negative examples. However, since NER is only the first step for our problem, blindly predicting the most common class will not work in this case, because we will not have any source/target/expression to predict our relationship on. Thus, we implemented another baseline model using the NLTK chunk labeling technique. (Loper & Bird, 2002) We first predict the part of speech for each token in our sentence, then we predict by mapping the part-of-speech tag to our labels. For example, if the part of speech tag of a word is a noun, then we predict it as a "B-TARGET", and if it is a verb then we predict it as a "B-EXPRESSION". With the predicted NER from part of speech tags, we then feed it as input into our relationship prediction baseline model, which is always predicting true for each relation tuple inputs.

We also compared the current state-of-the-art model by (Barnes et al., 2021) as a baseline. (Barnes et al., 2021) solve a similar problem using a neural graph parsing model, which is a reimplementation of the neural parser by (Dozat and Manning (2018) which was used by (Kurtz et al. 2020) for negation resolution. The parser learns to score each possible arc to then finally predict the output structure simply as a collection of all positively scored arcs.

## 4 Results and Discussion

### 4.1 Evaluation Metrics

The evaluation of the model performance on the task can be represented by the subtask performance on each subtask. Since the task of structured sentiment graph aggregation is trivial and deterministic, the evaluation metric of the task depends on the evaluation metric of the first two subtasks. The metrics take into consideration of existing metrics that have been used to compare the performance of structured sentiment analysis tasks.

The first opinion component extraction model can be evaluated using a simple $F_1$ score. To compare with state-of-art results, we evaluate the model using token-level $F_1$ scores for holders, targets, and expressions. A true positive is defined as the exact match of the predicted label and the true label of the token.

The second component relationship recognition task can be evaluated using the non-polar sentiment graph $F_1$. The metric measures how well the model can capture the relationship between the extracted components. True positive is defined as the exact match of the relationship between the holder, target, and the expression.

The combination of the two metrics above should measure the model's overall ability to extract structured sentiment from the text, in turn measuring its ability to extract opinions.

### 4.2 Evaluation Result

Overall, the proposed model outperforms the custom baseline as well as the existing state-of-the-art dependency graph parsing model in all common metrics for all datasets.

Table 2 shows the baseline model which demonstrates poor performance on all token F1 scores. It shows the improved performance in using the proposed model.

It is worth pointing out the proposed model does not predict the polarity of the sentiment expression thus polarity-related metrics are not compared.

Table 2 Average Token $F_1$ for all datasets

| Model | Holder $F_1$ | Target $F_1$ | Expression $F_1$ |
|---|---|---|---|
| Base | 0.49 | 0.45 | 0.48 |
| Final | 0.83 | 0.85 | 0.85 |

When comparing with the existing state-of-the-art models, we can see that the proposed model also demonstrates superior results.

Table 3 Token $F_1$ Per Dataset Comparing Models

| Dataset/ Model | Holder $F_1$ | Target $F_1$ | Expression $F_1$ |
|---|---|---|---|
| DS$_{unis}$ | - | - | - |
| Final | 0.50 | 0.77 | 0.75 |
| SOTA | 0.37 | 0.42 | 0.46 |
| MPQA | - | - | - |
| Final | 0.81 | 0.79 | 0.77 |
| SOTA | 0.46 | 0.51 | 0.48 |
| OpeNER | - | - | - |
| Final | 0.83 | 0.85 | 0.83 |
| SOTA | - | - | - |

And finally, the final proposed model's $F_1$ score for the relationship scoring is around 0.91 on average of three datasets. It has a precision of 0.93 and a recall of 0.96 on positive cases with a precision of 0.91 and a recall of 0.84 on negative cases. The baseline only achieved an $F_1$ score of 0.4 in comparison.

### 4.3 Analysis of Data Imbalance's Effect

From the original data analysis shown in Table 1, we understand that the data is imbalanced from the perspective of extracting the entities. But considering that negative relationship examples only existed in data that contained two or more target/expression pairs, the data imbalance problem also existed from the relationship labeling perspective.

For the relationship labeling, we create tuples for each pair of entity-expression and labeled positive for those who existed a relationship and negative for those who did not. From the final $F_1$ score of the proposed model, we can see the imbalance of the data has caused the model to score better on positive cases compared to negative cases.

We have applied data balancing methods like the one that was discussed in section 3.1 in an attempt to balance the distribution of the labels. However, the results were not promising so we kept the original distribution instead.

### 4.4 Analysis on the multiple targets evaluation

Another experiment that we have done is testing our model on different structured sentences to test for validity. We extracted the data where there is only one target within one sentence, and multiple targets within one sentence. Our model was able to handle both scenarios with a stable F1 score. Comparing the performance of (Barnes et al., 2021) with multiple targets, they hypothesize that predicting the full sentiment graph may have a larger effect on sentences with multiple targets. Table 4 shows the F1 score with multiple targets.

Table 4 Model $F_1$ for multi-target data

| Dataset/Model | Final $F_1$ | SOTA $F_1$ |
|---|---|---|
| $DS_{unis}$ | 0.81 | 0.43 |
| MPQA | 0.82 | 0.58 |
| OpeNER | 0.78 | - |

## 4.5 Analysis on the Effect of RoBERTa

Compared to the baseline model which used the NTLK library's NER capabilities, using transfer learning and the transformer architecture of RoBERTa improved the model's overall performance in extracting the entities of concern.

While the paper discussing the state-of-the-art model (Barnes et al., 2021) discussed the use of mBERT showed moderate gains, the newer and improved transfer learning enabled by the transformers demonstrated much more improvement compared to the baseline. The average improvement for each type of entity extraction is close to 2 times for NER tasks presented in this paper as demonstrated in Table 2.

## 5 Conclusion

In the paper, we have proposed a divide and conqueror approach using transformers and transfer learning in addressing the problem of structured sentiment analysis without polarity. The model has outperformed the state-of-the-art models on all three benchmark datasets available.

Although due to the scope of the paper, the experiments are only performed without the inclusion of polarity, we believe that polarity would only affect the potential benchmark metrics related to polarity, with the current metrics unchanged.

In the future, we would like to include polarity labels into the models through the NER component and access our model's capabilities in incorporating polarity into the equation. We would also like to explore the possibilities of potentially defining and measuring the opinion similarity to better understand the predicted outcome in comparison with our human expectations.